\documentclass{article}
\usepackage{spconf,amsmath,graphicx,amsthm,subfigure}
\usepackage{algorithm}
\usepackage{algorithmic}
\usepackage{xcolor,multirow,colortbl,overpic,tikz}
\usepackage{pgfplots}
\usepgfplotslibrary{patchplots}

\title{Isotropic Gaussian Splatting for Real-Time Radiance Field Rendering}
\name{Yuanhao Gong$^{1,2}$\thanks{This work is supported by National Natural Science Foundation of China (61907031) and Shenzhen Science and Technology Program (20231121165649002 and JCYJ20220818100005011),~ gong.ai@qq.com\\
	© 20XX IEEE. Personal use of this material is permitted. Permission from IEEE must be obtained for all other uses, in any current or future media, including reprinting/republishing this material for advertising or promotional purposes, creating new collective works, for resale or redistribution to servers or lists, or reuse of any copyrighted component of this work in other works}, Lantao Yu$^3$, Guanghui Yue$^{4}$}
\address{$^1$College of Electronics and Information Engineering, Shenzhen University, China\\ $^2$Guangdong Key Laboratory of Intelligent Information Processing, Shenzhen, China
	\\ $^3$Adobe Inc., San Jose, CA, USA\\
	$^4$School of Biomedical Engineering, Shenzhen University, Shenzhen, China}
\begin{document}
%
\maketitle
\begin{abstract}
The 3D Gaussian splatting method has drawn a lot of attention, thanks to its high performance in training and high quality of the rendered image. However, it uses anisotropic Gaussian kernels to represent the scene. Although such anisotropic kernels have advantages in representing the geometry, they lead to difficulties in terms of computation, such as splitting or merging two kernels. In this paper, we propose to use isotropic Gaussian kernels to avoid such difficulties in the computation, leading to a higher performance method. The experiments confirm that the proposed method is about {\bf 100X} faster without losing the geometry representation accuracy. The proposed method can be applied in a large range applications where the radiance field is needed, such as 3D reconstruction, view synthesis, and dynamic object modeling.  
\end{abstract}
\begin{keywords}
3DGS, isotropic, Gaussian splatting, radiance field, rendering
\end{keywords}
\section{Introduction}
\label{sec:intro}
Recently, the 3D Gaussian splatting method has gained significant attention in the field of 3D volumetric representation~\cite{Kerbl2023,Yu2023MipSplatting,guedon2023sugar}. This state-of-the-art technique has transformed the way complex geometry is approximated within a 3D space. By using 3D anisotropic Gaussian functions, the method is able to capture the intricate details of the underlying geometry with precision and accuracy. This advancement has unlocked new opportunities for advanced analysis and exploration in various applications, providing researchers and practitioners with valuable insights into the 3D volumetric representation.

\begin{figure}[!htb]
	\centering
	\includegraphics[width=0.7\linewidth]{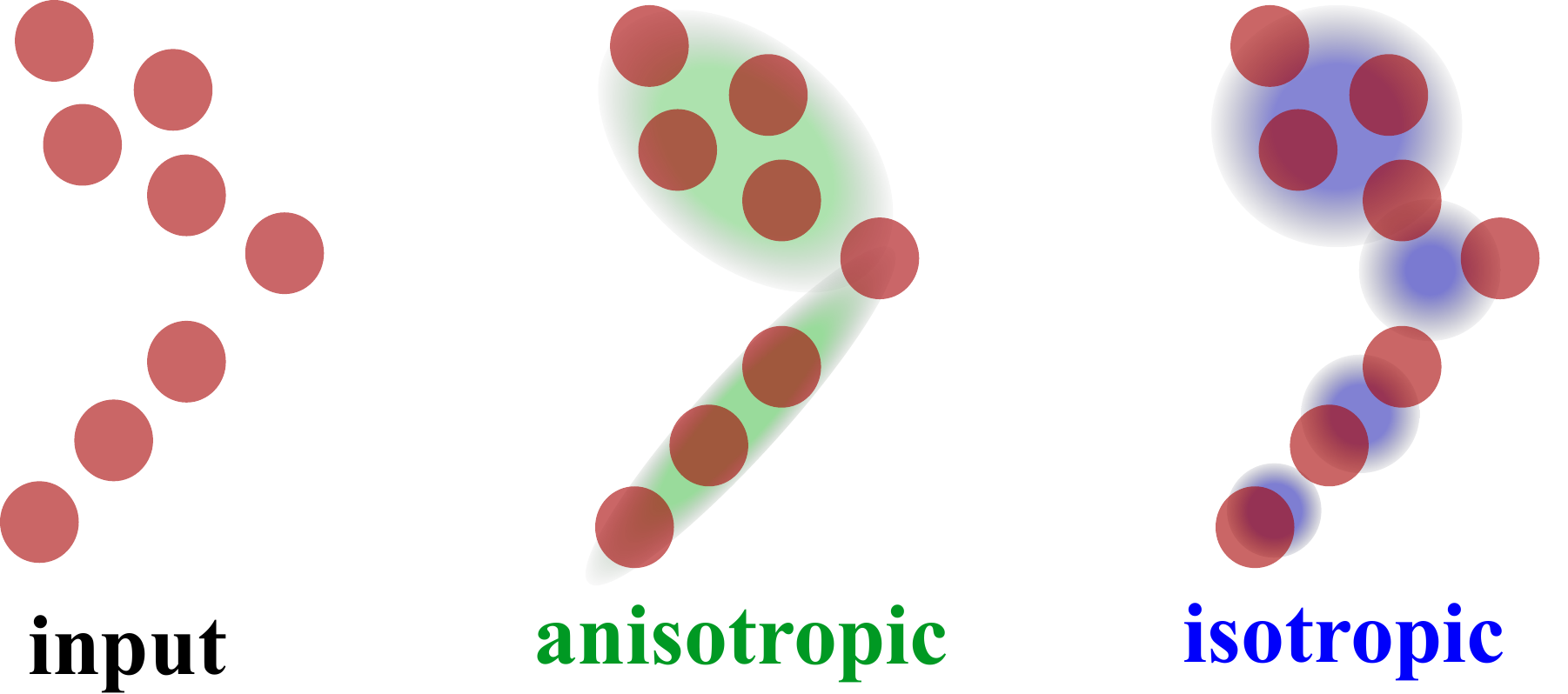}
	\caption{The input is indicated by the red dots. The green ellipses indicate the {\color{green}anisotropic} Gaussian kernels while the blue circles indicate the {\color{blue}\bf isotropic} Gaussian kernels. Although anisotropic Gaussian is more representative in terms of geometry, it leads to the computation difficulties. In contrast, isotropic Gaussian is more computational efficient.}
	\label{fig:illu}
\end{figure}
One of the major advantages of the 3D Gaussian splatting method is its exceptional ability to efficiently handle large-scale datasets. This method is specifically designed to process vast amounts of data in a relatively short amount of time, which makes it highly suitable for real-time applications. Its parallelizable nature allows for simultaneous processing of multiple data points, further enhancing its efficiency. By leveraging this technique, developers can effectively manage and analyze complex datasets without compromising on performance or accuracy. 

Additionally, the method allows for flexible parameter tuning. This adaptability is crucial in order to achieve the desired level of detail in the representation, making the 3D Gaussian splatting method an incredibly powerful and versatile tool in the field of 3D volumetric representation. With its ability to finely adjust parameters and accurately represent complex structures, the 3D Gaussian splatting method offers researchers unprecedented control and precision in their work, opening up new possibilities for advancements in various fields such as medical imaging~\cite{10230506}, computer graphics~\cite{Zhao}, and scientific visualization~\cite{Huang2023}. 
\subsection{Grid vs Particle}
In general, there are two ways to represent the 2D or 3D signals. One method is the grid-based representation, where the world is divided into a grid and each grid cell represents a specific region, for example the digital images and videos. Thanks to the regular sampling on the grid, the grid methods can effectively represent the low frequency and exchange long range (global) information. However, they have to use small grid size when dealing with the high frequency details (high resolution imaging), increasing the computation burden. 

Another approach is the particle representation. This approach entails depicting objects in the world as separate particles, each with unique properties. By embracing this particle representation framework, we can explore the intricate details and traits of the objects in question, enabling a more nuanced and comprehensive observation of the surrounding world. One typical example is the 3D Gaussian splatting~\cite{Kerbl2023}.

\subsection{Particle Representation}
Mathematically, any function $f(\vec{x})$ can be represented via convolution with a delta function~\cite{Reboux2012,Rossinelli2015}
\begin{equation}
	\label{eq:1}
	f(\vec{x})=\int\limits_{\vec{\tau}}f(\vec{\tau})\delta(\vec{x}-\vec{\tau})\mathrm{d}\vec{\tau}\,,
\end{equation} where $\vec{\tau},\vec{x}$ are spatial coordinates, $\delta(\cdot)$ is the Dirac delta function. In practice, the delta function is relaxed to a kernel with a compact support region. More specifically, the function can be approximated via
\begin{equation}
	\label{eq:2}
	f(\vec{x})\approx\hat{f}(\vec{x})\equiv\sum^K\limits_{||\vec{\tau}_k-\vec{x}||_2<D}f(\vec{\tau}_k)W(\vec{x}-\vec{\tau}_k, \theta)\,,
\end{equation} where $W(\cdot)\ge 0$ is a kernel (weight) function. It usually is normalized, $\sum W=1$. The $\theta$ is the kernel parameter. There are $K$ particles and $k$ is the particle index. The $\vec{\tau}_k$ is the center of the particle and $D$ indicates the support region size of the particle. $\hat{f}(\vec{x})$ is the reconstructed signal from the particles. The difference between them is the reconstruction error
\begin{equation}
	{\cal L}(f,\hat{f})=\frac{1}{2}\|f(\vec{x})-\hat{f}(\vec{x})\|^2_2\,.
\end{equation}
The gradients of the reconstruction error are
\begin{equation}
	\label{eq:3}
	\frac{\partial {\cal L}}{\partial \theta}=(\hat{f}-f)\sum f(\vec{\tau}_k)\frac{\partial W}{\partial \theta}\,,
\end{equation}
\begin{equation}
	\label{eq:32}
	\frac{\partial {\cal L}}{\partial \vec{\tau}}=-(\hat{f}-f)\sum f(\vec{\tau}_k)\frac{\partial W}{\partial \vec{\tau}}\,,
\end{equation} which are used to update $\theta$ and $\vec{\tau}$. 

Since the $W$ has local compact support, such methods are called particle methods, where each particle carries some properties, such as mass, curvature and temperature~\cite{Wang2022}.

\subsection{3D Gaussian Splatting}
The Gaussian splatting method is a widely used and effective technique in the field of computer graphics for rendering complex 3D scenes. It is employed to project three-dimensional points onto a two-dimensional image plane by utilizing a Gaussian kernel, which helps to achieve a visually pleasing and realistic representation. By employing this method, intricate details and nuances of the original 3D data can be accurately preserved and faithfully translated into a two-dimensional space, ensuring that crucial geometric and depth information is not lost in the process.

In the radiance field rendering, the color $C$ of a pixel can be computed via~\cite{Yifan2019}
\begin{equation}
	\label{eq:4}
	C=\sum_kc_k\alpha_k\prod_{j}^{k-1}(1-\alpha_j)\,,
\end{equation}where $c_k$ is the color of each point and $\alpha_k$ is the opacity. This is usually evaluated via ray-tracing, even the volume can be represented by points with radial basis functions~\cite{Xu2022}.

The 3D Gaussian splatting method uses Gaussian kernels
\begin{equation}
	\label{eq:5}
	G(\vec{x},\mu,\Sigma)=\exp^{-(\vec{x}-\vec{\mu})^T\Sigma^{-1}(\vec{x}-\vec{\mu})}\,,
\end{equation} where $\vec{\mu}$ is the center of the particle and $\Sigma$ is a symmetric non-negative covariant matrix. $\Sigma$ can be expressed as
\begin{equation}
	\label{eq:6}
	\Sigma=RSS^TR^T\,,
\end{equation} where $R$ is a rotation matrix and $S$ is a scaling matrix. This 3D function is then projected into 2D image space via a view transformation $V$ and a Jacobian of the affine approximation of the projection transformation $J$ by
\begin{equation}
	\label{eq:7}
	\Sigma^{2D}=JV\Sigma V^TJ^T\,.
\end{equation}

\subsection{Motivation and Contribution}
Although the anisotropic Gaussian kernel is more effective in representing the geometry (especially at edges), it leads to computation difficulties because of the orientation in $\Sigma$. In this paper, we propose to use scale-adaptive isotropic Gaussian kernels to acceleration the computation performance~\cite{Cheeseman2018}. Our contributions include
\begin{itemize}
	\item  we propose to use scale adaptive isotropic Gaussian kernels for signal representation.
	\item the isotropic Gaussian kernels have higher computational performance than the anisotropic ones.
	\item  several numerical experiments confirm the efficiency and effectiveness of the isotropic Gaussian kernels.
\end{itemize}
\section{Comparison between anisotropic and isotropic kernels}
\label{sec:igs}
In this section, we discuss the advantages and limitations of the anisotropic and isotropic Gaussian kernels. The isotropic Gaussian kernel with parameter $\mu$ and $\sigma^2$ is defined as
\begin{equation}
	\label{eq:8}
	g(\vec{x},\vec{\mu},\sigma^2)=\exp^{-\frac{(\vec{x}-\vec{\mu})^T(\vec{x}-\vec{\mu})}{\sigma^2}}\,.
\end{equation}
This kernel has less free parameters than the anisotropic one. Thus, with the same number of particles, this kernel shows lower rendering quality, as confirmed in~\cite{Kerbl2023}. In this paper, we show that more number of isotropic Gaussian kernels should be adopted to achieve the high quality and high performance.
\subsection{Parameters}
The $\Sigma$ in Eq.~\eqref{eq:5} is a $3\times 3$ matrix. Thanks to the symmetry property, it only has 6 free parameters. Taking the location $\vec{\mu}$ into account, there are 9 parameters in each particle. In contrast, the Eq.~\eqref{eq:8} has only 4 parameters to be estimated. As a result, the isotropic kernels are more computationally efficient. And smaller number of parameters benefits the optimization process.
\begin{figure}[!tb]
	\centering
	\subfigure[aniso vs iso kernels]{
		\includegraphics[width=0.4\linewidth]{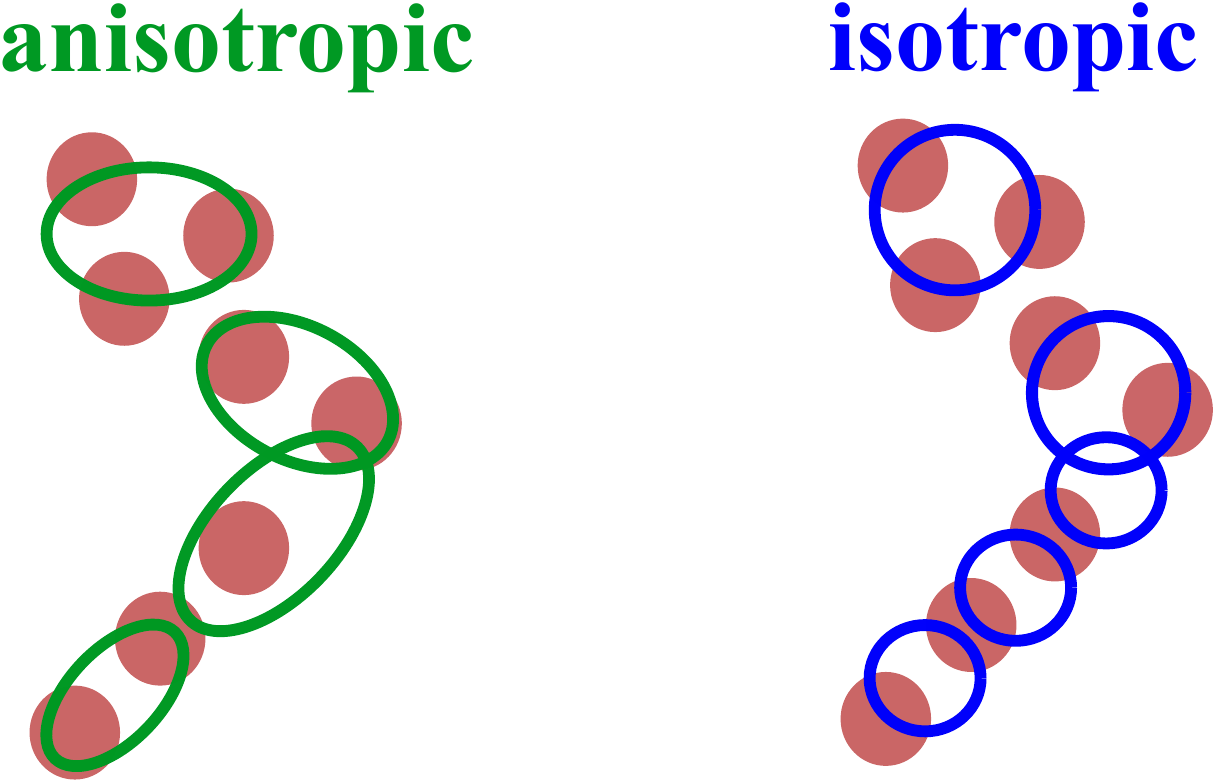}}~~~~
	\subfigure[adaptive iso kernel in 2D]{	\includegraphics[width=0.4\linewidth]{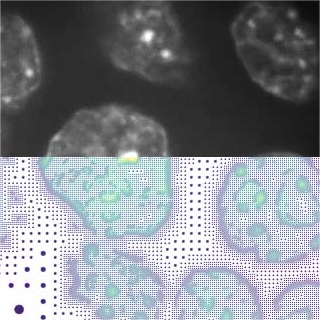}}
	\caption{(a) The input is indicated by the red dots. Green and blue lines indicate anisotropic and isotropic kernels respectively. (b) The image can be represented by size-adaptive isotropic particles. The top half is the pixel-based image and the bottom half is the particle representation. The dot size indicates the Gaussian kernel size. Although the sharp edges use more particles, the isotropic kernels lead to higher computational performance.}
	\label{fig:2}
\end{figure}
\subsection{Probability}
In the field of image analysis, it is widely recognized that both 2D and 3D images have a distinct feature known as sparse edges~\cite{gong:gdp}. These edges, which define the boundaries and outlines of objects within an image, are characterized by a relatively low density of pixels or points. The notion of sparsity in edges is well-established and acknowledged in the areas of image processing and computer vision. 

In other words, the most regions in an image are flat, indicating the isotropic property of the image. Therefore, even using the anisotropic kernels, most of the kernels are still isotropic at most regions thanks to the input data. The anisotropic kernels only appear at edges, which are sparse.

\subsection{Interaction between Kernels}
Despite the statistics, the interaction between anisotropic kernels is more computational expensive than the isotropic ones. As shown in the Fig.~\ref{fig:2}(a), the interaction between anisotropic kernels rely on the location, scale and orientation. Indeed, the orientation leads to more accurate representation, especially at the edges. However, such orientation also causes the difficulty to distribute the input onto the particles, especially in the 3D case where the orientation becomes nontrivial. 

In contrast, the isotropic kernels only rely on location and scale. And their interaction is much simpler and  more computationally efficient, leading to a higher performance. At sharp edges, we can use more small size particles to represent the edge, as shown in Fig.~\ref{fig:2}(b). And the particle size can depend on the local curvature~\cite{Wang2022}.
\begin{table} 
	\centering \caption{Compare anisotropic and isotropic Gaussian kernels.}
	\begin{tabular}{c|c|c}
		\hline
		&aniso&iso\\
		\hline
		parameters $\downarrow$&9$\times$&4$\times$\\
		probability $\uparrow$ &low &high\\
		interaction $\downarrow$ &high &low\\
		view dependency &yes&no\\
		multiscale$\downarrow$ &complex & easy\\
		\hline
	\end{tabular}
	\label{table}
\end{table}
\subsection{View Dependency}
Another difference between the anisotropic and isotropic kernels is the dependency on the view point. With different view points, the anisotropic kernels might show very different shapes. The orientation of the view point and the orientation of the kernel make the coupled system more complex. 

In contrast, the isotropic kernels do not have this issue. Its size only depends on the distance to the view point and its independent from the view orientation. Such property significantly reduces the computation complexity. 
\subsection{Multiscale}
Although the anisotropic Gaussian kernels can be represented in multi-scale, their orientation makes the scale decoupling challenge. In contrast, the isotropic Gaussian kernels can be easily merged, divided, deleted or added, according to their scale settings. One example is shown in Fig.~\ref{fig:2}(b). These differences are summarized in Table~\ref{table}.
\section{Our Method}
We represent the signal via weighted isotropic Gaussian kernels. More specifically, the reconstructed signal is
\begin{equation}
	\hat{f}(\vec{x})=\sum\limits_{||\vec{\tau}_k-\vec{x}||_2<D} A_kg(\vec{x}-\vec{\mu}_k,\sigma_k)\,,
\end{equation} where $A_k$ is the property (such as color or opacity), $\vec{\mu}_k$ is the center of the particle and $\sigma_k$ controls the scale.

Our method has two stages. The first initialization stage is based on the tree structure. The second stage is the optimization process to reduce the following reconstruction error
\begin{equation}
	{\cal L}=(1-\lambda)\|f(\vec{x})-\hat{f}(\vec{x})\|_1+\lambda\cdot {\mathrm{SSIM}}(f(\vec{x}),\hat{f}(\vec{x}))\,,
\end{equation} where $\lambda=0.2$ by default.
\subsection{Initial with Tree Structure}
\begin{figure}[!htb]
	\centering
	\subfigure[original]{\includegraphics[width=0.45\linewidth]{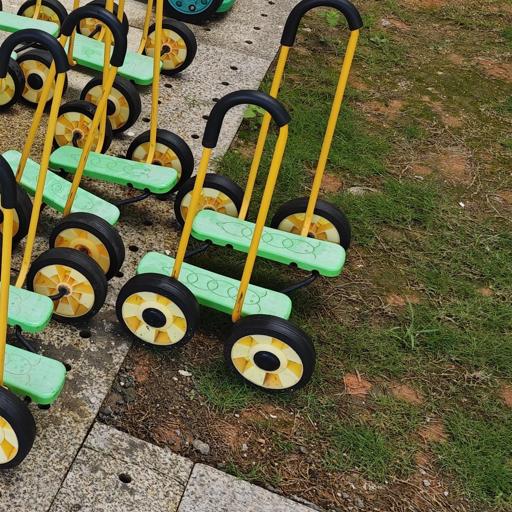}}~~
	\subfigure[2D quad tree]{\includegraphics[width=0.45\linewidth]{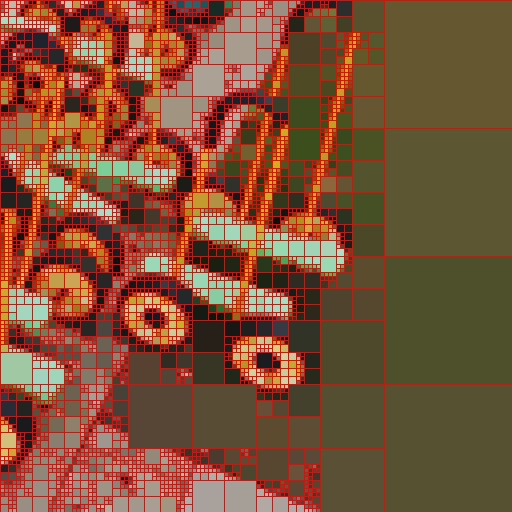}}
	\caption{The tree structure is adopted to initialize and organize the particles. Each cell contains one or more particles.}
	\label{fig:tree}
\end{figure}

Instead of using the random initialization as in the original Gaussian splatting method, we use a QuadTree and Octree to initialize and manage the particles in 2D and 3D signals, respectively. One example is shown in Fig.~\ref{fig:tree}. 

\begin{figure}[!htb]
	\centering
	\subfigure[initial state]{\includegraphics[width=0.4\linewidth]{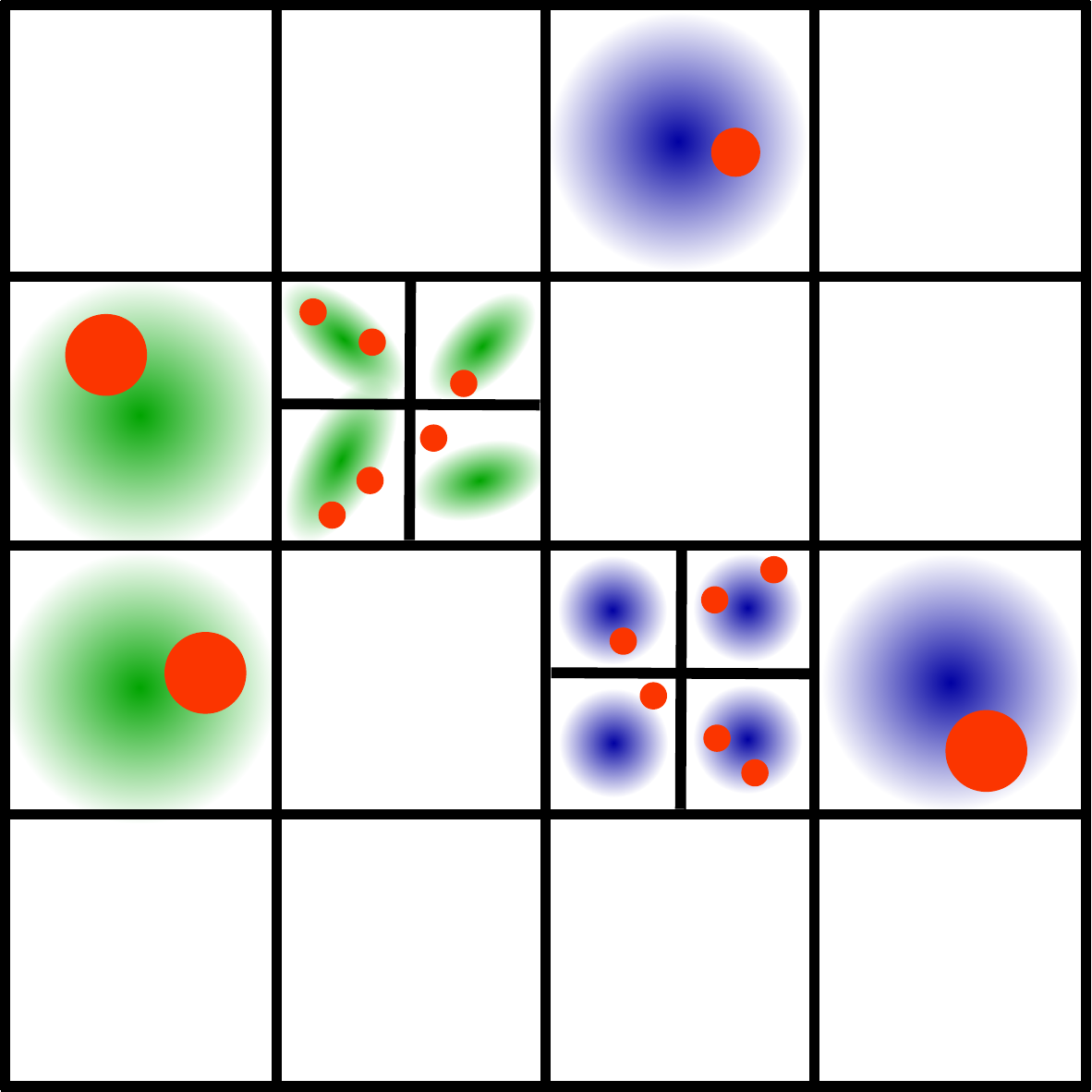}}
	\begin{minipage}{0.05\linewidth}
		\centering
		\vspace{-3.3cm}
		$\Rightarrow $
	\end{minipage}
	\subfigure[state during optimizing]{\includegraphics[width=0.4\linewidth]{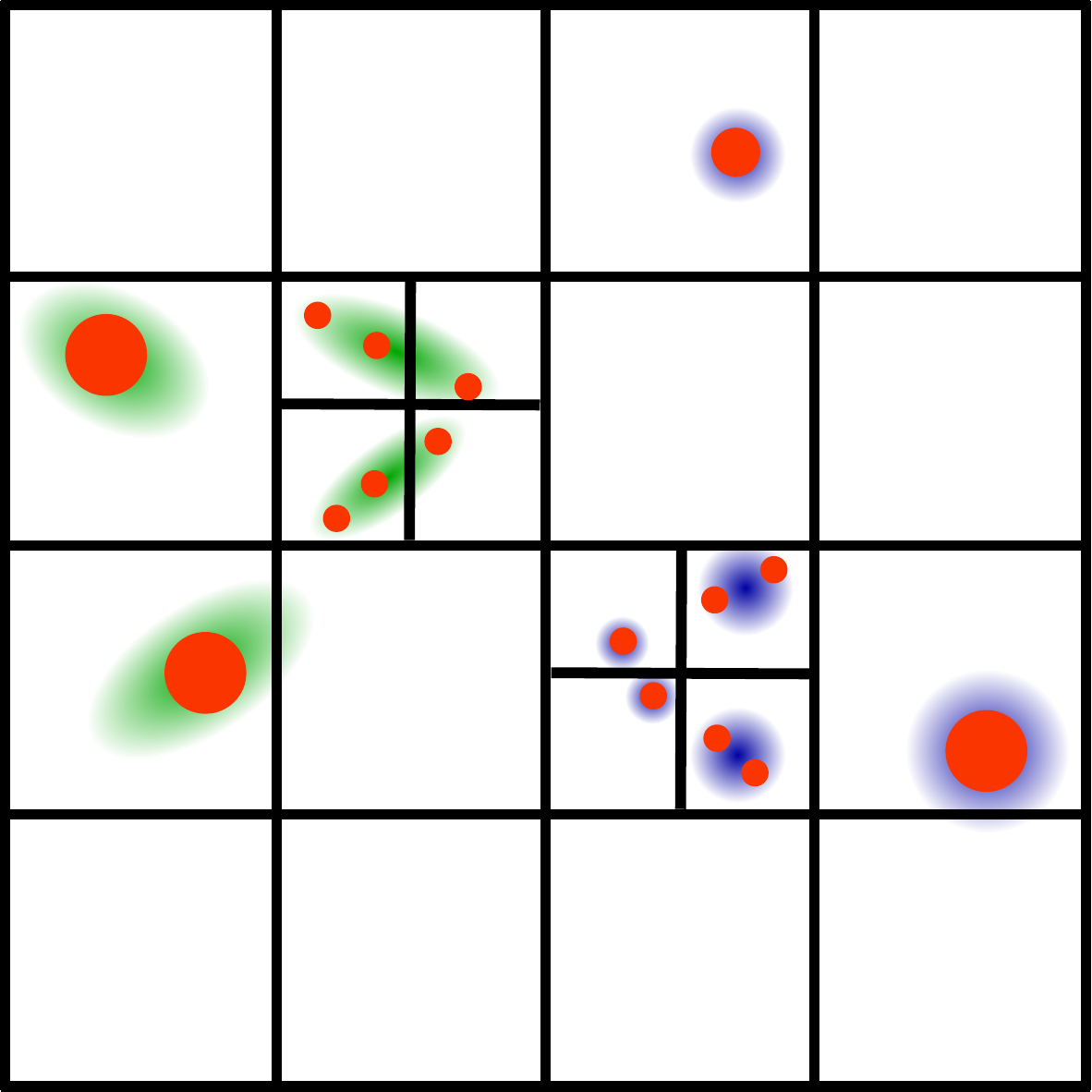}}
	\caption{The anisotropic and isotropic kernels in a QuadTree structure. The red is the input data. Green and blue indicate anisotropic and isotropic kernels, respectively. (a) is the initial state. (b) is a state during optimization.}
	\label{fig:tree2}
\end{figure}

Each leaf node in the tree is a cell that contain one or more particles that carry the colors and opacity in that region. The anisotropic and isotropic kernels are illustrated in Fig.~\ref{fig:tree2}. We initialize the color $A_k$ as the average in the cell. The scale $\sigma$ is initialized as half of the cell width. The idea is illustrated in Fig.~\ref{fig:tree2}.

\begin{table} 
	\centering \caption{Running time in minutes.}
	\begin{tabular}{c|c|c|c|c}
		\hline
		setting&$(15,10^3)$&$(50,10^3)$&$(15,2*10^3)$&$(50,2*10^3)$\\
		\hline
		aniso&69m&73m&121m&128m\\
		iso  &0.12m &0.13m&0.25m&0.27m\\
		\hline
	\end{tabular}
	\label{table2}
\end{table}
\subsection{Optimization}
In general, there are two ways to optimize the loss function. One is the classical back propagation. The other is the forward-only evolution algorithms. The particles can be deleted if $A_k$ is smaller than a threshold. They can also be merged or split, as shown in \cite{Kerbl2023}.

\subsection{Experiments}
\begin{figure}
	\centering
	\subfigure[$D=15, K=10^3$]{\includegraphics[width=0.23\linewidth]{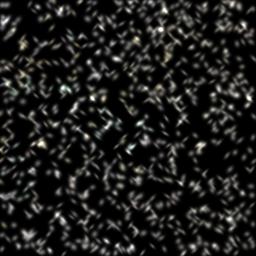}~\includegraphics[width=0.23\linewidth]{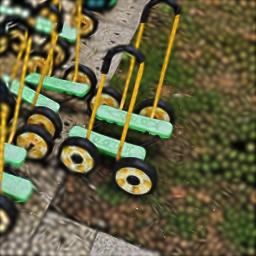}}
	\subfigure[$D=50, K=10^3$]{\includegraphics[width=0.23\linewidth]{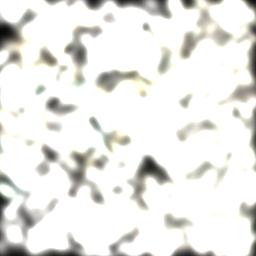}~\includegraphics[width=0.23\linewidth]{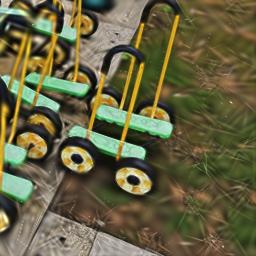}}
		\subfigure[$D=15, K=2\times10^3$]{\includegraphics[width=0.23\linewidth]{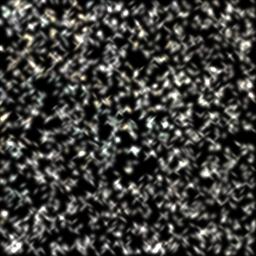}~\includegraphics[width=0.23\linewidth]{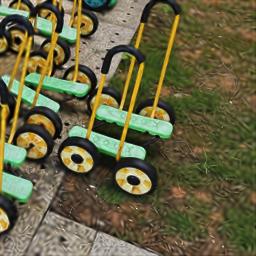}}
		\subfigure[$D=50, K=2\times10^3$]{\includegraphics[width=0.23\linewidth]{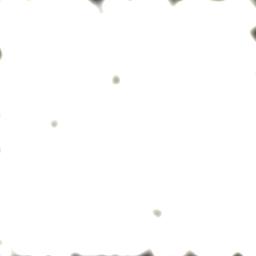}~\includegraphics[width=0.23\linewidth]{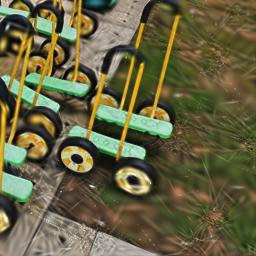}}
		\caption{Anisotropic Gaussian kernels with different parameter settings. The left in each panel is the random initialization. The right is the resulting image after 2000 epochs.}
		\label{fig:ex1}
\end{figure}

\begin{figure}
	\centering
	\subfigure[Tree Init (s=3), $K=10^3$]{\includegraphics[width=0.23\linewidth]{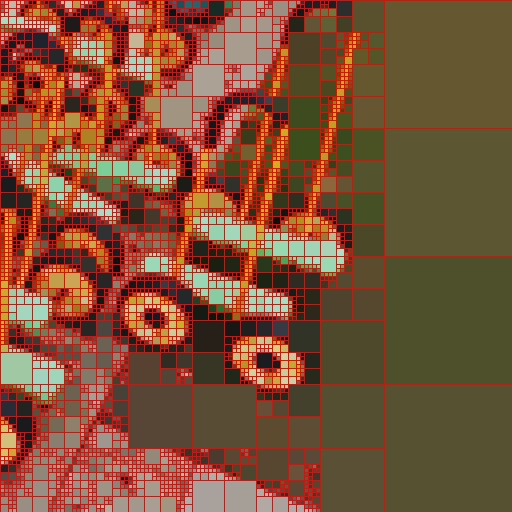}~\includegraphics[width=0.23\linewidth]{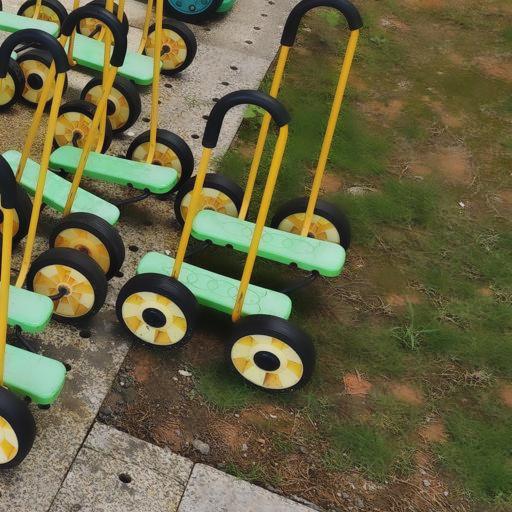}}
	\subfigure[Tree Init (s=7), $ K=10^3$]{\includegraphics[width=0.23\linewidth]{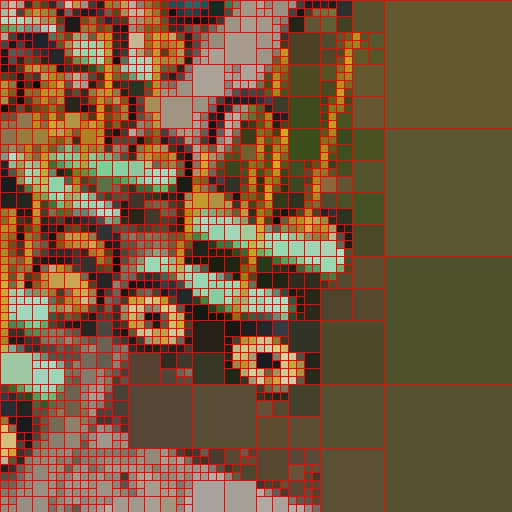}~\includegraphics[width=0.23\linewidth]{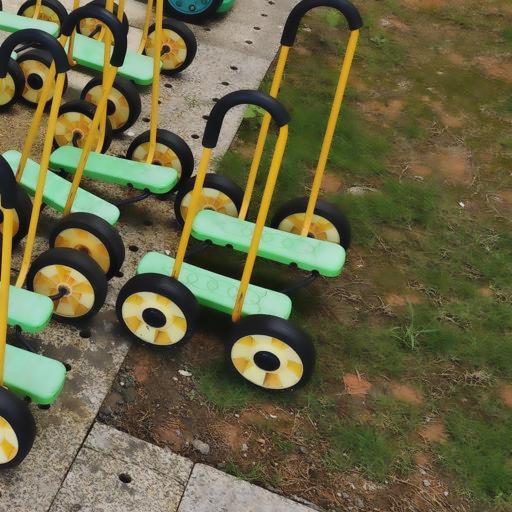}}
	\subfigure[Tree Init (s=3), $K=2\times10^3$]{\includegraphics[width=0.23\linewidth]{img/d3.jpg}~\includegraphics[width=0.23\linewidth]{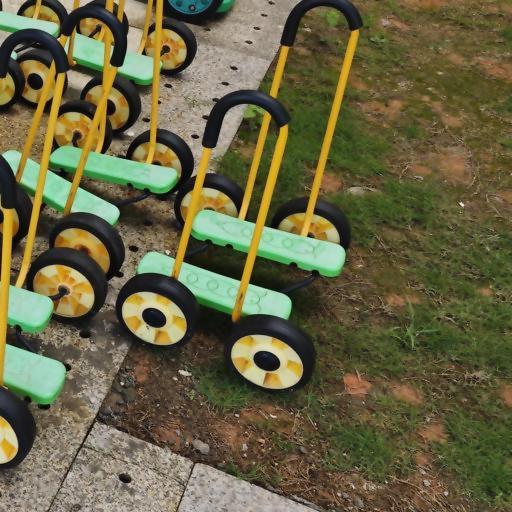}}
	\subfigure[Tree Init (s=7), $ K=2\times10^3$]{\includegraphics[width=0.23\linewidth]{img/d7.jpg}~\includegraphics[width=0.23\linewidth]{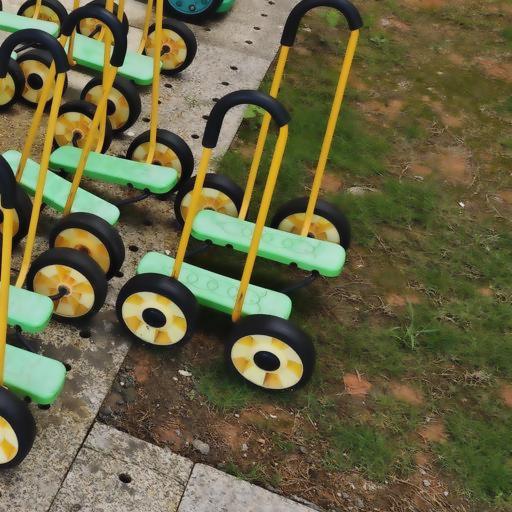}}
	\caption{Isotropic kernels with different parameter settings. The left in each panel is the tree-based initialization.}
	\label{fig:ex2}
\end{figure}

One example for the anisotropic and isotropic kernels is shown in Figs.~\ref{fig:ex1} and \ref{fig:ex2}, respectively. Clearly, our method can achieve higher quality with less artifacts. Moreover, the proposed method is much faster in the training process, as shown in Table~\ref{table2}. Our method is about 100 times faster.

\section{Conclusion}
In this paper, we propose to use isotropic Gaussian kernel for geometry representation, which has higher computation performance than the anisotropic case. The proposed method can be applied in various image processing and 3D rendering tasks~\cite{Xu2023,Yu2019,gong:cf,Yu2022a,Tang2023a,Yu2020,10230506,GONG2019329,Yin2019,Tang2023}. 


\bibliographystyle{../IEEEbib}
\bibliography{../../IP}

\begin{thebibliography}{10}

\bibitem{Kerbl2023}
Bernhard Kerbl, Georgios Kopanas, Thomas Leimkuehler, and George Drettakis,
\newblock ``3d gaussian splatting for real-time radiance field rendering,''
\newblock {\em ACM Trans. Graph.}, vol. 42, no. 4, jul 2023.

\bibitem{Yu2023MipSplatting}
Zehao Yu, Anpei Chen, Binbin Huang, Torsten Sattler, and Andreas Geiger,
\newblock ``Mip-splatting: Alias-free 3d gaussian splatting,''
\newblock {\em arXiv:2311.16493}, 2023.

\bibitem{guedon2023sugar}
Antoine Gu{\'e}don and Vincent Lepetit,
\newblock ``Sugar: Surface-aligned gaussian splatting for efficient 3d mesh
  reconstruction and high-quality mesh rendering,''
\newblock {\em arXiv preprint arXiv:2311.12775}, 2023.

\bibitem{10230506}
Yuanhao Gong, Wanlin Huang, and Wenhui Wu,
\newblock ``Removing scattered light in biomedical images,''
\newblock in {\em 2023 IEEE 20th International Symposium on Biomedical Imaging
  (ISBI)}, 2023, pp. 1--5.

\bibitem{Zhao}
Zhibo Zhao, Wenming Tang, and Yuanhao Gong,
\newblock ``Curvature-driven multi-stream network for feature-preserving mesh
  denoising,''
\newblock {\em Computer Graphics Forum}, vol. n/a, no. n/a, pp. e14993.

\bibitem{Huang2023}
Wanlin Huang, Wenhui Wu, and Yuanhao Gong,
\newblock ``3d hand bones and tissue estimation from a single 2d x-ray image
  via a two-stream deep neural network,''
\newblock in {\em 2023 IEEE 20th International Symposium on Biomedical Imaging
  (ISBI)}, 2023, pp. 1--5.

\bibitem{Reboux2012}
Sylvain Reboux, Birte Schrader, and Ivo~F. Sbalzarini,
\newblock ``A self-organizing lagrangian particle method for
  adaptive-resolution advection–diffusion simulations,''
\newblock {\em Journal of Computational Physics}, vol. 231, no. 9, pp.
  3623--3646, 2012.

\bibitem{Rossinelli2015}
Diego Rossinelli, Babak Hejazialhosseini, Wim {van Rees}, Mattia Gazzola,
  Michael Bergdorf, and Petros Koumoutsakos,
\newblock ``Mrag-i2d: Multi-resolution adapted grids for remeshed vortex
  methods on multicore architectures,''
\newblock {\em Journal of Computational Physics}, vol. 288, pp. 1--18, 2015.

\bibitem{Wang2022}
Cheng Wang, Wanli Wang, Shucheng Pan, and Fuyu Zhao,
\newblock ``A local curvature based adaptive particle level set method,''
\newblock {\em Journal of Scientific Computing}, vol. 91, no. 1, pp. 3, 2022.

\bibitem{Yifan2019}
Wang Yifan, Felice Serena, Shihao Wu, Cengiz \"{O}ztireli, and Olga
  Sorkine-Hornung,
\newblock ``Differentiable surface splatting for point-based geometry
  processing,''
\newblock {\em ACM Trans. Graph.}, vol. 38, no. 6, nov 2019.

\bibitem{Xu2022}
Qiangeng Xu, Zexiang Xu, Julien Philip, Sai Bi, Zhixin Shu, Kalyan Sunkavalli,
  and Ulrich Neumann,
\newblock ``Point-nerf: Point-based neural radiance fields,''
\newblock New Orleans, LA, USA, 2022, pp. 5428--5438, IEEE.

\bibitem{Cheeseman2018}
Bevan~L. Cheeseman, Ulrik Günther, Krzysztof Gonciarz, Mateusz Susik, and
  Ivo~F. Sbalzarini,
\newblock ``Adaptive particle representation of fluorescence microscopy
  images,''
\newblock {\em Nature Communications}, vol. 9, no. 1, pp. 5160, 2018.

\bibitem{gong:gdp}
Y.~Gong and I.F. Sbalzarini,
\newblock ``A natural-scene gradient distribution prior and its application in
  light-microscopy image processing,''
\newblock {\em Selected Topics in Signal Processing, IEEE Journal of}, vol. 10,
  no. 1, pp. 99--114, Feb 2016.

\bibitem{Xu2023}
Meng Xu, Zhihuang Zhang, Yuanhao Gong, and Stefan Poslad,
\newblock ``Regression-based camera pose estimation through multi-level local
  features and global features,''
\newblock {\em Sensors}, vol. 23, no. 8, 2023.

\bibitem{Yu2019}
Lantao Yu and Michael~T. Orchard,
\newblock ``Single image interpolation exploiting semi-local similarity,''
\newblock Brighton, UK, 2019, pp. 1722--1726, IEEE.

\bibitem{gong:cf}
Yuanhao Gong and Ivo~F. Sbalzarini,
\newblock ``Curvature filters efficiently reduce certain variational
  energies,''
\newblock {\em IEEE Transactions on Image Processing}, vol. 26, no. 4, pp.
  1786--1798, April 2017.

\bibitem{Yu2022a}
Lantao Yu, Dehong Liu, Hassan Mansour, and Petros~T. Boufounos,
\newblock ``Fast and high-quality blind multi-spectral image pansharpening,''
\newblock {\em IEEE Transactions on Geoscience and Remote Sensing}, vol. 60,
  pp. 1--17, 2022.

\bibitem{Tang2023a}
Wenming Tang, Yuanhao Gong, and Guoping Qiu,
\newblock ``Feature preserving 3d mesh denoising with a dense local graph
  neural network,''
\newblock vol. 233, pp. 103710, 2023.

\bibitem{Yu2020}
Lantao Yu, Dehong Liu, Hassan Mansour, Petros~T. Boufounos, and Yanting Ma,
\newblock ``Blind multi-spectral image pan-sharpening,''
\newblock in {\em ICASSP 2020 - 2020 IEEE International Conference on
  Acoustics, Speech and Signal Processing (ICASSP)}, 2020, pp. 1429--1433.

\bibitem{GONG2019329}
Yuanhao Gong and Orcun Goksel,
\newblock ``Weighted mean curvature,''
\newblock {\em Signal Processing}, vol. 164, pp. 329 -- 339, 2019.

\bibitem{Yin2019}
H.~{Yin}, Y.~{Gong}, and G.~{Qiu},
\newblock ``Side window filtering,''
\newblock in {\em Proc. IEEE/CVF Conf. Computer Vision and Pattern Recognition
  (CVPR)}, 2019, pp. 8750--8758.

\bibitem{Tang2023}
Wenming Tang, Zewei Lin, and Yuanhao Gong,
\newblock ``Gc-net: An unsupervised network for gaussian curvature optimization
  on images,''
\newblock {\em Journal of Signal Processing Systems}, vol. 95, no. 1, pp.
  77--88, 2023.

\end{thebibliography}

\end{document}